\documentclass[pdflatex,sn-mathphys-num]{sn-jnl}
\usepackage{url}

\usepackage{graphicx}%
\usepackage{multirow}%
\usepackage{amsmath,amssymb,amsfonts}%
\usepackage{amsthm}%
\usepackage{mathrsfs}%
\usepackage[title]{appendix}%
\usepackage{xcolor}%
\usepackage{textcomp}%
\usepackage{manyfoot}%
\usepackage{booktabs}%
\usepackage{algorithm}%
\usepackage{algorithmicx}%
\usepackage{algpseudocode}%
\usepackage{listings}%
\usepackage{caption}

\theoremstyle{thmstyleone}%
\theoremstyle{thmstyletwo}%

\theoremstyle{thmstylethree}%

\begin{document}

\title[Federated Learning for Cross-Modality Medical Image Segmentation via Augmentation-Driven Generalization]{Federated Learning for Cross-Modality Medical Image Segmentation via Augmentation-Driven Generalization}


\author*[1]{\fnm{Sachin} \sur{Dudda Nagaraju}}\email{sachin.d.nagaraju@ntnu.no}
\author[1]{\fnm{Ashkan} \sur{Moradi}}\email{ashkan.moradi@ntnu.no}
\author[1]{\fnm{Bendik Skarre} \sur{Abrahamsen}}\email{bendik.s.abrahamsen@ntnu.no}
\author[1,2]{\fnm{Mattijs} \sur{Elschot}}\email{mattijs.elschot@ntnu.no}

\affil*[1]{\orgdiv{Department of Circulation and Medical Imaging}, \orgname{Norwegian University of Science and Technology}, \orgaddress{\city{Trondheim}, \country{Norway}}}
\affil[2]{\orgdiv{Central Staff}, \orgname{St. Olavs Hospital, Trondheim University Hospital}, \orgaddress{\city{Trondheim}, \country{Norway}}}


\abstract{
\textbf{Purpose:}
Developing generalizable medical image segmentation models is challenging because imaging data are distributed across institutions and differ in modality and acquisition protocol. Federated learning (FL) enables collaborative training without centralizing raw medical images, but cross-modality domain shifts between computed tomography (CT) and magnetic resonance imaging (MRI) can substantially reduce model performance. This study investigates augmentation-driven cross-modality FL for abdominal organ and whole-heart segmentation.

\textbf{Methods:}
We evaluate convolution-based spatial augmentation, frequency-domain argumentation, domain-specific normalization, and global intensity nonlinear (GIN) augmentation for multimodal segmentation. Abdominal organ segmentation and whole-heart segmentation are first evaluated using a 2D U-Net framework. For whole-heart segmentation, we additionally perform native 3D experiments using a self-configuring nnU-Net architecture on the CARE-WHS 2026 dataset, enabling evaluation of whether the observed cross-modality FL behavior persists when moving from slice-based 2D segmentation to volumetric 3D segmentation.

\textbf{Results:}
GIN provides the most consistent cross-modality performance among the evaluated approaches in the original 2D experiments. For pancreas segmentation, the Dice similarity coefficient (DSC) improved from 0.073 to 0.437 when CT data were incorporated through federated cross-modality training. In 3D whole-heart segmentation, FedGIN improved mean DSC over FedAvg from 0.8696 to 0.8901 on the unseen CT center and from 0.7160 to 0.7956 on the unseen MRI center. Relative to centralized GIN training, FedGIN retained 92.4\% of performance on unseen CT data and achieved comparable performance on unseen MRI data (0.7956 versus 0.7937).

\textbf{Conclusions:}
GIN-based augmentation improves cross-modality federated segmentation by introducing intensity variation while preserving anatomical structure.
}

\keywords{Federated Learning, Medical Image Segmentation, Cross-Modality Generalization, Unpaired Multimodal Data, Multi-institutional Collaboration.}



\maketitle

\section{Introduction}

\label{sec:introduction}
\begin{figure}[!t]
    \centering
    \includegraphics[width=0.8\textwidth]{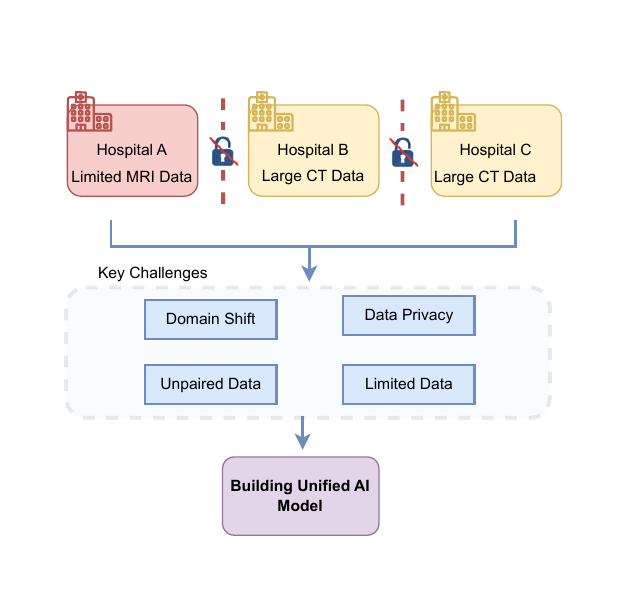}
    \caption{Key challenges in federated multi-site medical image segmentation.}
    \label{fig:challenges}
\end{figure}

Modern healthcare systems are increasingly relying on computer-assisted medical image analysis to support clinical decision-making, such as organ segmentation in treatment planning and diagnosis. However, deploying robust and generalizable segmentation models across varied healthcare institutions continues to pose significant challenges\cite{ogut2025artificial,rao2025multimodal}. Developing modality-agnostic segmentation models that can operate effectively across different imaging protocols (e.g., CT and MRI) is essential for widespread clinical adoption \cite{litjens2017survey}. Such models would enable healthcare institutions to collaborate and benefit from collective knowledge regardless of their available imaging equipment, improve model generalization by leveraging complementary information from multiple modalities, and reduce the need for modality-specific model development and maintenance \cite{zhou2019review}. However, achieving this goal faces substantial obstacles. As shown in Fig.~\ref{fig:challenges}, building unified and generalizable segmentation models for multi-site medical imaging faces three core challenges. First, privacy regulations like HIPAA and GDPR prohibit the sharing of patient data across institutions. This limitation calls for alternative strategies like distributed learning~\cite{mehrtabar2025ethical}. Second, technical heterogeneity across modalities and acquisition parameters creates substantial domain shifts; for example, when segmenting abdominal organs across institutions, CT and MRI scans of the same organ exhibit drastically different intensity characteristics, spatial resolution, noise patterns, and tissue contrast~\cite{zhou2022generalizable,stacke2020measuring}. Additionally, each center typically has limited data for each modality, which may cause conventional models to fail at producing meaningful segmentations. Therefore, developing an approach that efficiently leverages the combination of such limited data from different modalities is desirable. Third, paired multimodal data is rarely available across institutions, since patients typically undergo either CT or MRI at a given site, making retrospective cross-modal pairing across sites impractical. These challenges are further exacerbated by the uneven distribution of data across institutions
~\cite{groger2025review}.

Federated Learning (FL)~\cite{mcmahan2017communication} offers a privacy-preserving way for institutions to train models collaboratively by sharing model updates instead of raw data. Although FL facilitates multi-site training, it does not inherently mitigate inter-institutional domain shifts. Several approaches have been proposed in the medical imaging literature to address cross-domain generalization, including domain adaptation techniques using adversarial training~\cite{javanmardi2018domain} and domain-specific batch normalization (DSBN) ~\cite{zhou2022generalizable} reduce distribution gaps but increase complexity. Data augmentation techniques such as FedDG~\cite{liu2021feddg} and FDA~\cite{li2023frequency} enhance style diversity by applying frequency-domain perturbations, modifying the Fourier amplitude to generate varied image appearances. Random convolution-based augmentation~\cite{ouyang2022causality,scholz2025random,choi2023progressive} has shown promise by applying spatial filters that alter low-level appearance characteristics such as noise and contrast while preserving anatomical structure. Despite these advances, a critical gap remains: existing comparisons of augmentation strategies have been conducted in centralized settings where all modalities are available during training. In this work, we consider a realistic yet underexplored FL scenario where each client holds data from a single modality (e.g., CT or MRI), reflecting common real-world conditions where institutions typically specialize in one imaging modality. Under this constraint, models must generalize to unseen modalities without direct exposure during local training. This single-modality-per-client setting fundamentally changes which augmentation strategies can be effective, as techniques designed for centralized multimodal training may not directly translate to such federated settings. However, no systematic evaluation has addressed this gap. 

In our previous work, we introduced FedGIN~\cite{nagaraju2025fedgin}, a lightweight intensity-based augmentation technique for federated organ segmentation. By applying random convolutional transformations during local training, FedGIN matched centralized performance while preserving privacy. It was particularly effective for organs that are inherently difficult to segment, such as the pancreas and gallbladder. 

Motivated by recent advances in augmentation-based approaches for federated medical imaging~\cite{nagaraju2025fedgin}, this work systematically investigates which strategies best enable cross-modality generalization under realistic federated constraints. Specifically, we adapt and extend GIN (Global Intensity Nonlinear) augmentation, originally proposed for single-domain generalization~\cite{ouyang2022causality}, to the federated cross-modality setting. We then comprehensively evaluate three categories of generalization strategies: spatial-domain augmentation (random convolution-based methods), frequency-domain augmentation (Fourier amplitude augmentation), and network-level adaptation (domain-specific batch normalization). Through extensive experiments on binary and multi-class segmentation tasks, we demonstrate that spatial-domain augmentation via GIN consistently outperforms alternative approaches by directly modifying intensity distributions to simulate cross-modality variations while preserving anatomical structure.

\noindent The main contributions of this work are:
\begin{itemize}

\item We propose FedGIN, an FL framework that integrates on-the-fly GIN augmentation for cross-modality medical image segmentation, enabling robust generalization across CT and MRI without sharing raw patient data.

\item We present a systematic analysis of augmentation and adaptation strategies for federated multimodal medical image segmentation, showing that GIN-based spatial augmentation provides more consistent cross-modality performance than frequency-domain and domain-specific normalization approaches.

\item We conduct extensive experiments on multi-site abdominal organ segmentation and whole-heart segmentation, demonstrating cross-modality generalization without requiring paired CT--MRI data or modifications to the underlying segmentation network.

\item We further validate the federated cross-modality findings using native 3D whole-heart segmentation with a self-configuring nnU-Net architecture, complementing the primary 2D experiments conducted using the widely adopted U-Net architecture. We additionally introduce a bidirectional holdout evaluation using unseen CT and unseen MRI centers, revealing that the performance retained under federated training differs across modalities and centers.

\end{itemize}

The rest of the paper is structured as follows: Section~\ref{rw} discusses related work, Section~\ref{meth} outlines our methodology, Section~\ref{exp} covers experiments, results and analysis in detail, and Section~\ref{conc} concludes with insights.

\section{Related Works}
\label{rw}

This section reviews the literature across two interconnected areas: FL for medical imaging and domain generalization techniques for cross-modality segmentation.

\subsection{Federated Learning in Medical Imaging}

FL has emerged as a distributed learning paradigm that enables collaborative model training without centralizing sensitive patient data. The FedAvg algorithm~\cite{mcmahan2017communication} established the foundation by averaging locally trained model weights, enabling distributed optimization while keeping data on-premise. Subsequent works focused on unique challenges of medical imaging federations. FedProx~\cite{li2020federated} introduced a proximal regularization term to handle statistical heterogeneity arising from non-IID data distributions across hospitals. SCAFFOLD~\cite{karimireddy2020scaffold} proposed variance reduction techniques to correct client drift, improving convergence stability in heterogeneous settings. In medical imaging, FL has been successfully applied to several tasks. These applications include brain tumor segmentation~\cite{pati2021federated}, cancer detection~\cite{moradi2025optimizing}, COVID-19 detection from medical images~\cite{nguyen2021federated}, and other medical diagnostic tasks~\cite{sandhu2023medical}. Recent advances have explored personalization strategies, communication efficiency, and differential privacy integration to address practical deployment constraints\cite{liu2024recent,moradi2025beyond}. Many medical FL studies consider settings in which participating clients operate on the same imaging modality, while cross-modality heterogeneity across clients has received comparatively less attention. When institutions contribute data from different scanners or modalities (e.g., CT versus MRI), domain-specific differences can lead to divergent local model updates, reducing the effectiveness of standard federated aggregation~\cite{rehman2023federated}. This form of multimodal heterogeneity, particularly when cross-modality data are unpaired across institutions, remains comparatively underexplored~\cite{thrasher2025multimodal}.

\subsection{Domain Generalization for Cross-Modality Medical Image Segmentation}

Domain generalization (DG) aims to train models that generalize to unseen target domains, a critical capability for cross-modality medical imaging. Existing approaches can be broadly categorized into three paradigms: network-level adaptation, frequency-domain augmentation, and spatial-domain augmentation.

Network-level methods modify architectural components to handle domain variations. Zhou et al.~\cite{zhou2022generalizable} employed domain-specific batch normalization (DSBN), in which separate normalization statistics are maintained for different domains while the remaining network parameters are shared. Such approaches can effectively accommodate domain-specific feature distributions, although their complexity increases as additional domains are introduced. Frequency-domain approaches manipulate Fourier representations to simulate domain variations. FreeSDG~\cite{li2023frequency} and RaffeSDG~\cite{li2024raffesdg} introduce frequency-domain perturbations to increase appearance diversity while preserving structural information. FedDG~\cite{liu2021feddg} extends frequency-based augmentation to federated settings by sharing amplitude-spectrum information across clients, enabling domain diversification without exchanging raw images. However, frequency-domain transformations primarily modify spectral appearance characteristics and may not fully capture the complex nonlinear intensity differences that occur between modalities such as CT and MRI. Spatial-domain augmentation directly perturbs image intensities through random transformations. Random convolution-based methods~\cite{ouyang2022causality,choi2023progressive} apply randomly initialized convolutional filters with varying or fixed kernel sizes to generate diverse appearance and texture variations while preserving anatomical structure. These approaches have demonstrated promising generalization in single-source settings, but their effectiveness for cross-modality medical image segmentation remains comparatively underexplored, particularly in federated scenarios where different imaging modalities are distributed across clients~\cite{li2025federated}. Cross-modal medical image segmentation has also been investigated using image-translation approaches that synthesize target-modality images for training~\cite{zhang2018translating} or jointly optimize appearance adaptation and segmentation~\cite{chen2020unsupervised}. However, image-translation approaches may require paired or co-registered data, additional synthesis networks, or assumptions regarding the relationship between source and target modality distributions, which can complicate their use in multi-institutional federated settings.

Existing approaches therefore present several limitations for federated cross-modality segmentation. Network-level adaptation may become increasingly complex as the number of domains grows, while frequency-domain techniques may not fully represent nonlinear modality-specific intensity variations. Spatial augmentation provides an alternative means of increasing appearance diversity without modifying the segmentation architecture, but its effectiveness under federated cross-modality heterogeneity remains comparatively underexplored. We address these limitations by adapting and systematically evaluating augmentation and adaptation strategies for cross-modality federated learning without requiring paired CT--MRI observations.

\begin{figure*}[!t]
\centering
\includegraphics[width=\textwidth]{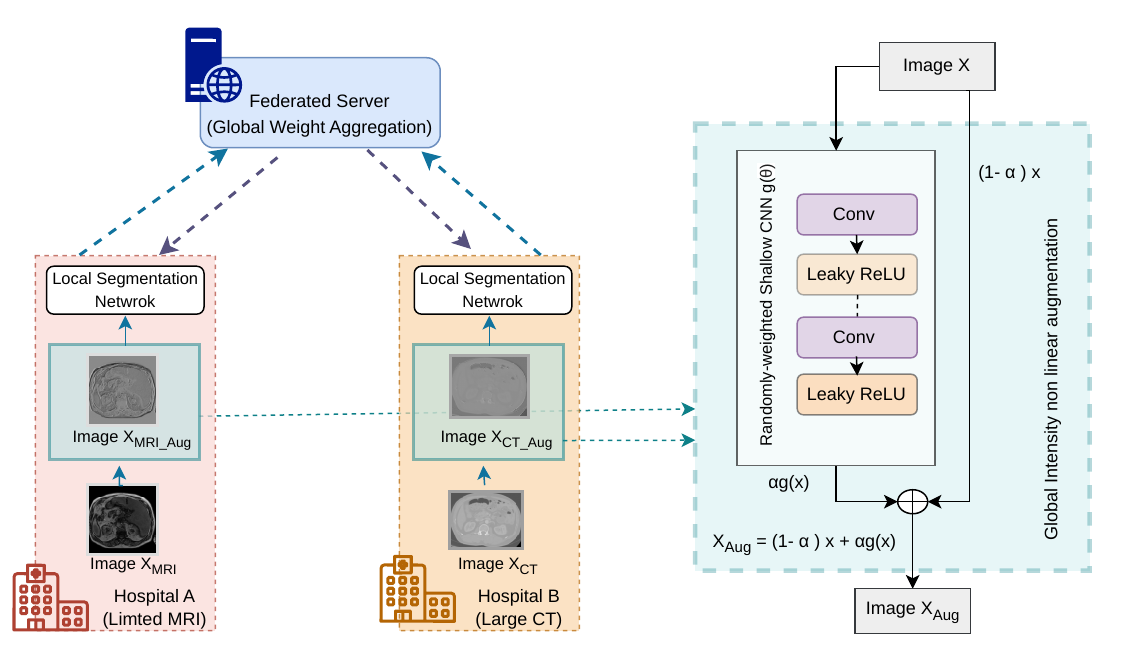}
\caption{Overview of the proposed FedGIN framework showing server-coordinated federated training and local GIN augmentation. The augmentation module applies randomly initialized convolutional layers to generate non-linear intensity transformations while preserving spatial correspondence. The transformed input is interpolated with the original input as
$X_{\mathrm{Aug}}=(1-\alpha)X+\alpha g(X)$, where $\alpha\in[0,1]$ controls the augmentation strength.}
\label{fig:gin+}
\end{figure*}

\section{Methodology}
\label{meth}

In this section, we describe our FL framework for cross-modality medical image segmentation. We first formalize the problem, then introduce the random convolution-based GIN augmentation method, and finally describe the federated training procedure.

\subsection{Problem Formulation}

Consider an FL scenario involving $K$ participating clients, where each client
$k \in \{1,2,\ldots,K\}$ maintains a local dataset
$\mathcal{D}_k=\{(x_i^k,y_i^k)\}_{i=1}^{N_k}$ containing $N_k$ medical imaging samples $x_i^k$ and their corresponding segmentation masks $y_i^k$. Each client contains data from a single imaging modality,
$m_k \in \{\mathrm{CT},\mathrm{MRI}\}$, while different clients may contain different modalities. No paired CT--MRI observations from the same subject are required across clients. The objective is to collaboratively learn a global segmentation model $f_{\theta}$ from these heterogeneous single-modality datasets such that the resulting model generalizes effectively across both CT and MRI data.

\subsection{Global Intensity Non-linear Augmentation}

The GIN module is a key component of our approach, enabling cross-modality generalization within the FL framework.

\subsubsection{Design}

The core idea behind GIN is that while imaging modalities differ in intensity and contrast, the underlying anatomical structures remain consistent. By introducing random intensity transformations during training, GIN encourages the model to focus on learning anatomical features that are consistent across modalities, rather than  modality-specific intensity patterns.

 For the 2D experiments GIN operates on the image while preserving its spatial dimensions, i.e.,
$g(\cdot): \mathbb{R}^{C \times H \times W} \rightarrow
\mathbb{R}^{C \times H \times W}$, where $H$ and $W$ denote the spatial
dimensions and $C$ represents the number of image channels
($C=1$ for the grayscale medical images used in this study).
The convolutional transformation is spatially equivariant and does not
introduce geometric deformation, thereby maintaining the spatial
correspondence of anatomical structures.

\subsubsection{GIN Architecture}
The GIN module instantiates intensity transformations through a shallow
randomly initialized convolutional network. The network consists of
$L=4$ convolutional layers with randomly selected kernel sizes
$k \in \{1,3\}$ and randomly sampled weights and biases.
Each convolution uses stride 1 and padding
$p=\lfloor k/2 \rfloor$ to preserve the spatial dimensions of the input.
The random convolutional parameters are not optimized together with the
segmentation network. LeakyReLU activation with a negative slope of
$0.01$ is applied after the first $L-1$ convolutional layers, while the
final layer remains linear. The resulting transformations modify image
intensity and local appearance without introducing geometric
transformations of the anatomical structures.

\subsubsection{Transformation Formulation}

The GIN transformation pipeline is illustrated in Fig.~\ref{fig:gin+} (right). For an input image $x$, the transformation is defined as:
\begin{equation}
g(x) =
x_{\text{mix}}
\frac{\|x\|_F}
{\|x_{\text{mix}}\|_F+\epsilon},
\qquad \epsilon=10^{-5},
\label{eq:gin_complete}
\end{equation}
where
\begin{equation}
x_{\text{mix}}
=
\alpha\, g^{\text{Net}}(x)
+
(1-\alpha)x,
\label{eq:mixing}
\end{equation}
Here, $g^{\text{Net}}(\cdot)$ denotes the random convolutional
transformation, $\alpha \sim \mathcal{U}(0,1)$ controls the interpolation
between the transformed and original inputs, and $\|\cdot\|_F$ denotes
the Frobenius norm.

The random network $g^{\text{Net}}$ consists of $L$ sequential layers:
\begin{equation}
g^{\text{Net}}(x) =
h_L \circ h_{L-1} \circ \cdots \circ h_1(x),
\end{equation}
where each layer $h_\ell$ applies:
\begin{equation}
h_\ell(z) =
\begin{cases}
\operatorname{Conv}(z;\mathbf{W}_\ell,\mathbf{b}_\ell),
& \ell=L, \\[2mm]
\operatorname{LeakyReLU}
\left(
\operatorname{Conv}(z;\mathbf{W}_\ell,\mathbf{b}_\ell);
0.01
\right),
& \ell<L.
\end{cases}
\end{equation}
The parameters $\mathbf{W}_\ell$ and $\mathbf{b}_\ell$ are randomly
sampled and are not optimized together with the segmentation network.
They are resampled for each augmentation call during training.

For the supporting 3D experiments, the same augmentation
principle is applied directly to volumetric inputs by replacing the
two-dimensional random convolutions with their three-dimensional
counterparts, using kernel sizes $1\times1\times1$ and
$3\times3\times3$. The detailed 3D experimental configuration is
described in Section~\ref{exp}.

The transformation therefore proceeds through three stages:
(1) random intensity mapping through $g^{\text{Net}}$,
(2) stochastic interpolation with the original input, and
(3) Frobenius-norm rescaling to preserve the overall input magnitude.

\subsection{Federated Training Framework}

\subsubsection{Framework Overview}

Our framework integrates on-the-fly GIN augmentation with federated aggregation, based on the principle that segmentation models should rely primarily on modality-robust anatomical features rather than modality-specific intensity cues. As illustrated in Fig.~\ref{fig:gin+}, a central server coordinates training across multiple clients containing different imaging modalities (e.g., MRI and CT), while each client applies GIN augmentation locally during training. The augmentation introduces random nonlinear intensity and appearance variations, thereby broadening the local training distribution and reducing reliance on modality-specific characteristics. This strategy aims to improve cross-modality generalization across heterogeneous institutional datasets.

\begin{algorithm}[t]
\caption{Federated Learning with On-the-Fly GIN Augmentation}
\label{alg:fedgin}
\small
\begin{algorithmic}[1]

\Require Number of clients $K$, communication rounds $T$,
local epochs $E$, GIN depth $L$
\Ensure Trained global model parameters $\theta^{(T)}$

\State Initialize global model parameters $\theta^{(0)}$

\For{$t=0$ to $T-1$}

    \State Broadcast $\theta^{(t)}$ to all participating clients

    \ForAll{clients $k\in\{1,\ldots,K\}$ \textbf{in parallel}}

        \State $\theta_k \gets \theta^{(t)}$

        \For{$e=1$ to $E$}

            \ForAll{mini-batches $(X_b,Y_b)\subset\mathcal{D}_k$}

                \State Sample kernel sizes
                $k_\ell\in\{1,3\}$, $\ell=1,\ldots,L$

                \State Sample random GIN parameters
                $\mathbf{W}_\ell,\mathbf{b}_\ell
                \sim\mathcal{N}(0,\mathbf{I})$

                \State Sample interpolation coefficient
                $\alpha\sim\mathcal{U}(0,1)$

                \State $\widetilde{X}_b \gets g(X_b)$
                \Comment{On-the-fly GIN augmentation}

                \State $\mathcal{L}\gets
                \mathcal{L}_{\mathrm{seg}}
                \bigl(f_{\theta_k}(\widetilde{X}_b),Y_b\bigr)$

                \State Update $\theta_k$ using the local optimizer
                to minimize $\mathcal{L}$

            \EndFor
        \EndFor

        \State $\theta_k^{(t+1)}\gets\theta_k$

    \EndFor

    \State $N\gets\sum_{k=1}^{K}N_k$

    \State $\displaystyle
    \theta^{(t+1)}
    \gets
    \sum_{k=1}^{K}
    \frac{N_k}{N}
    \theta_k^{(t+1)}$

\EndFor

\State \Return $\theta^{(T)}$

\end{algorithmic}
\end{algorithm}

\subsubsection{Local Training with On-the-fly GIN Augmentation}

At each communication round $t$, client $k$ receives the global model parameters $\theta^{(t)}$ and performs local training. For each training sample $(x,y)\in\mathcal{D}_k$, GIN augmentation is applied on-the-fly. The input passes through the GIN module, and the augmented output is directly fed to the segmentation network. Since the GIN parameters are randomly sampled for each augmentation call, successive forward passes generate different intensity and appearance transformations, exposing the model to a broader range of input variations throughout training.

\subsubsection{Global Model Aggregation}

After local training, each client transmits its updated model parameters
$\theta_k^{(t+1)}$ to the server. The client models are aggregated using
FedAvg~\cite{mcmahan2017communication}:
\begin{equation}
\theta^{(t+1)}
=
\sum_{k=1}^{K}
\frac{N_k}{N}
\theta_k^{(t+1)},
\label{eq:fedavg}
\end{equation}
where $N_k$ denotes the number of training samples at client $k$ and
$N=\sum_{k=1}^{K}N_k$ is the total number of training samples across all
participating clients. Thus, each client's contribution to the global
model is weighted according to its local dataset size. The resulting
global model $\theta^{(t+1)}$ is then broadcast to the clients for the
next communication round.

\subsection{Training Overview}

Algorithm~\ref{alg:fedgin} summarizes the complete federated training
procedure with on-the-fly GIN augmentation. The main advantages of the
framework are: (1) compatibility with existing segmentation architectures
without requiring modifications to the segmentation network, and
(2) the ability to perform cross-modality federated training without
requiring paired CT--MRI data. These characteristics make the framework
suitable for multi-institutional medical imaging settings where modality
availability varies across participating sites. 

\section{Experiments and Results}
\label{exp}
This section presents the experimental evaluation of our proposed FL framework for cross-modality medical image segmentation, covering two use cases: binary abdominal organ segmentation and multiclass whole heart segmentation. We first describe the general setup, then present results for each use case, and conclude with a broader discussion.

\subsection{General Setup}

\subsubsection{Implementation Details}

For the 2D experiments, we use a 2D U-Net backbone~\cite{nagaraju2025fedgin} implemented in PyTorch 2.6.0, with federated training orchestrated using Flower 1.20.0~\cite{beutel2020flower}. All experiments are conducted on an NVIDIA A40 GPU with 48 GB of memory. Optimization uses AdamW \cite{loshchilov2017decoupled} with a learning rate of $5\!\times\!10^{-4}$ and weight decay of $1\!\times\!10^{-4}$, combined with a focal and Dice loss function and a plateau learning rate scheduler with automatic mixed precision. All federated experiments run for $T = 100$ communication rounds with $E = 1$ local epoch per round.

For primary 2D experiments, we adopt a slice-based segmentation approach for both abdominal organ segmentation and whole-heart segmentation. This choice is motivated by three practical considerations: slice thickness varies significantly (1--5~mm) across scanners and modalities, making volumetric 3D convolution inconsistent; 2D slicing increases the number of training samples available per federated client; CT and MRI tend to have more comparable image characteristics in the axial plane than through-plane, making 2D cross-modality alignment more tractable. All 3D volumes were sliced along the axial plane to generate 2D training inputs, and final evaluation metrics were computed as 3D Dice scores after reconstructing the predicted slices back into original volumes.

In addition to this 2D pipeline, we conduct a complementary native 3D experiment for whole-heart segmentation using a self-configuring nnU-Net architecture \cite{isensee2021nnu}, described separately in Section~\ref{sec:3d-whs} together with its implementation details, as its data handling differs fundamentally from the slice-based approach used elsewhere in this study.

\subsubsection{Data Preprocessing}


All volumes are reoriented to RAS (Right-Anterior-Superior) orientation and resampled to a consistent in-plane resolution of $1.5 \times 1.5$ mm (256$\times$256 pixels). Through-plane (z-axis) spacing is preserved from the original acquisition, except for high-resolution volumes where original spacing $< 2.5$mm is resampled to 3mm. Linear interpolation is used for images and nearest-neighbor interpolation for segmentation masks. Intensities are normalized using per-volume z-score normalization before axial slice extraction.

\subsubsection{Baseline Methods}

We compare our proposed approach against representative methods from three categories, all adapted to the federated cross-modality segmentation setting. As a non-augmentation baseline, FedAvg~\cite{mcmahan2017communication} applies standard federated averaging without any domain generalization strategy. For network-level adaptation, we evaluate DSBN~\cite{zhou2022generalizable}, which employs domain-specific batch normalization with separate statistics per modality. For augmentation-based methods, we evaluate frequency-domain approaches including FMAug~\cite{li2023frequency} and RaffeSDG~\cite{li2024raffesdg}, as well as spatial-domain methods including ProRandConv~\cite{choi2023progressive} and RC-Unet~\cite{scholz2025random}. All methods were originally designed for single-source domain generalization in centralized settings and are adapted here to the federated framework using FedAvg for aggregation.

\begin{table}[h]
\centering
\caption{Distribution of 3D volumes in training, validation, and testing sets for abdominal organ segmentation.}
\label{tab:organ_data}
\begin{tabular}{lcccccc}    
\hline
\multirow{2}{*}{\textbf{Organ}} & \multicolumn{2}{c}{\textbf{Training}} & \multicolumn{2}{c}{\textbf{Validation}} & \multicolumn{2}{c}{\textbf{Testing}} \\
 & MRI & CT & MRI & CT & MRI & CT \\
\hline
Liver        & 80 & 704 & 20 & 177 & 60 & 60 \\
Kidneys      & 84 & 662 & 22 & 166 & 60 & 60 \\
Gall Bladder & 65 & 510 & 17 & 128 & 60 & 60 \\
Spleen       & 81 & 687 & 21 & 172 & 60 & 60 \\
Pancreas     & 76 & 631 & 19 & 158 & 60 & 60 \\
\hline
\end{tabular}
\end{table}

\subsection{Use case 1: Abdominal Organ Segmentation}

\subsubsection{Dataset}

We evaluate abdominal organ segmentation using two public 3D medical imaging datasets. The TotalSegmentator dataset~\cite{wasserthal2023totalsegmentator}, which contains both CT and MRI volumes, is used for training and validation. The AMOS dataset~\cite{ji2022amos} serves as an independent and unseen test set to evaluate the generalization of the modality-agnostic model. We focus on five abdominal organs: liver, kidneys, spleen, pancreas, and gallbladder. For federated experiments, data is partitioned across clients such that each client possesses only a single modality (either CT or MRI). Table~\ref{tab:organ_data} summarizes the data distribution across training and validation sets.

The TotalSegmentator dataset exhibits typical medical imaging heterogeneity, with voxel spacing of approximately $1.5 \times 1.5 \times 1.5$~mm for CT and $1.4 \times 1.4 \times 6.0$~mm for MRI, while AMOS2020 has spacing of approximately $0.8 \times 0.8 \times 5.0$~mm, reflecting real-world variability across institutions.

\subsubsection{Experimental Setup and Motivation}

We design two complementary experiments for this use case. The first investigates whether a hospital with scarce MRI data can improve model performance by collaborating with CT-rich institutions through FL, a scenario that is common in practice, where annotated data is typically available for a single modality at each site. We establish a baseline using 20 MRI cases from TotalSegmentator and progressively augment it with CT data from federated client (20, 40, and 100 cases), evaluating on the AMOS2020 MRI test set.

The second experiment evaluates cross-modality generalization using the full dataset from both modalities. We assess whether FL can achieve bidirectional generalization comparable to centralized training. For each of the five organs, we train and evaluate the following configurations: (1) MRI-only, trained exclusively on MRI and tested on both modalities; (2) CT-only, trained exclusively on CT and tested on both modalities; (3) Centralized (CT+MRI), trained on pooled data with and without augmentation; and (4) FL, where K=2 clients collaborate through FL, each possessing data from only one modality, and the aggregated global model is tested on both CT and MRI.

\begin{table}[h]
\centering
\caption{FedGIN improvement over MRI-only baseline (20 cases) on AMOS2020 MRI test set with 100 CT cases added.}
\label{tab:organ_limited_summary}
\begin{tabular}{lccc}
\hline
\textbf{Organ} & \textbf{Baseline} & \textbf{FedGIN} & \textbf{Gain} \\
\hline
Liver & 0.851 & 0.770 & -9.4\% \\
Kidneys & 0.740 & 0.833 & +12.6\% \\
Spleen & 0.662 & 0.773 & +17.1\% \\
Gall Bladder & 0.160 & 0.403 & +151.9\% \\
Pancreas & 0.073 & 0.437 & +498.6\% \\
\hline
\end{tabular}
\end{table}

\begin{table*}[t]
\centering
\caption{Cross-Modality Segmentation Performance (DSC) on AMOS2020 Test Set for Abdominal Organ Segmentation. All models trained on TotalSegmentator dataset. Best results in \textbf{bold}.}
\label{tab:organ_results}
\resizebox{\textwidth}{!}{%
\begin{tabular}{l|cc|cc|cc|cc|cc}
\hline
\multirow{2}{*}{\textbf{Method}} & \multicolumn{2}{c|}{\textbf{Liver}} & \multicolumn{2}{c|}{\textbf{Kidneys}} & \multicolumn{2}{c|}{\textbf{Spleen}} & \multicolumn{2}{c|}{\textbf{Gall Bladder}} & \multicolumn{2}{c}{\textbf{Pancreas}} \\
& MRI & CT & MRI & CT & MRI & CT & MRI & CT & MRI & CT \\
\hline
\multicolumn{11}{l}{\textit{Single Modality Local Training}} \\
\hline
MRI-only & 0.8511 & 0.2632 & 0.8901 & 0.0440 & 0.7731 & 0.0372 & 0.3221 & 0.0128 & 0.1229 & 0.0019 \\
CT-only & 0.3909 & 0.8967 & 0.7708 & 0.8711 & 0.6511 & 0.8809 & 0.0760 & 0.5169 & 0.2512 & 0.6411 \\
\hline
\hline
\multicolumn{11}{l}{\textit{Centralized (CT+MRI) Multi-Modality Training}} \\
\hline
No augmentation & 0.9157 & 0.9132 & 0.8925 & 0.8807 & 0.8812 & 0.9050 & 0.0229 & 0.4701 & 0.6109 & 0.6130 \\
DSBN \cite{zhou2022generalizable} & 0.7811 & 0.9205 & 0.7405 & 0.8371 & 0.1100 & 0.0229 & 0.0760 & 0.5693 & 0.1498 & 0.4158 \\
FMAug \cite{li2023frequency} & 0.7807 & 0.8181 & 0.6541 & 0.1312 & 0.6481 & 0.5813 & 0.3447 & 0.0270 & 0.0840 & 0.0041 \\
RaffeSDG \cite{li2024raffesdg} & 0.8830 & 0.8332 & 0.7511 & 0.7461 & 0.5697 & 0.8012 & 0.3329 & 0.5818 & 0.5111 & 0.5153 \\
ProRandConv \cite{choi2023progressive} & 0.7061 & 0.9055 & 0.8610 & 0.8698 & 0.8126 & 0.8813 & 0.4064 & 0.5521 & 0.6113 & 0.6097 \\
RC-Unet \cite{scholz2025random} & 0.8859 & 0.9233 & 0.8882 & 0.8678 & 0.8604 & 0.8742 & 0.5405 & 0.4971 & 0.6156 & 0.6218 \\
\textbf{GIN} & \textbf{0.9187} & \textbf{0.9301} & \textbf{0.9112} & \textbf{0.9016} & \textbf{0.8910} & \textbf{0.9108} & \textbf{0.5934} & \textbf{0.6124} & \textbf{0.6564} & \textbf{0.6804} \\
\hline
\hline
\multicolumn{11}{l}{\textit{Federated Learning (CT Client + MRI Client)}} \\
\hline
FedAvg \cite{mcmahan2017communication} & 0.7645 & 0.8551 & 0.7569 & 0.8764 & 0.7537 & 0.8636 & 0.0861 & 0.4486 & 0.2309 & 0.6211 \\
DSBN & 0.1395 & 0.8781 & 0.0117 & 0.9014 & 0.0233 & 0.7609 & 0.0023 & 0.5445 & 0.0021 & 0.6006 \\
FMAug & 0.7162 & 0.8288 & 0.0523 & 0.0184 & 0.7220 & 0.6915 & 0.1568 & 0.0164 & 0.0003 & 0.0667 \\
RaffeSDG & 0.6324 & 0.8910 & 0.1248 & 0.6284 & 0.5308 & 0.7922 & 0.1007 & 0.4964 & 0.0989 & 0.5007 \\
ProRandConv & 0.7059 & 0.9221 & 0.8506 & 0.8597 & 0.6938 & 0.8946 & 0.4227 & 0.5989 & 0.4474 & 0.6378 \\
RC-Unet & 0.5953 & 0.9071 & 0.4193 & 0.8787 & 0.7681 & 0.8829 & 0.0324 & 0.5697 & 0.0045 & 0.6054 \\
\textbf{FedGIN} & \textbf{0.8530} & \textbf{0.9340} & \textbf{0.8972} & \textbf{0.9037} & \textbf{0.8824} & \textbf{0.8905} & \textbf{0.5532} & \textbf{0.6002} & \textbf{0.6567} & \textbf{0.6909} \\
\hline
\end{tabular}%
}
\end{table*}

\begin{figure*}[p]
    \centering
    \includegraphics[width=\textwidth, height=\textheight, keepaspectratio]{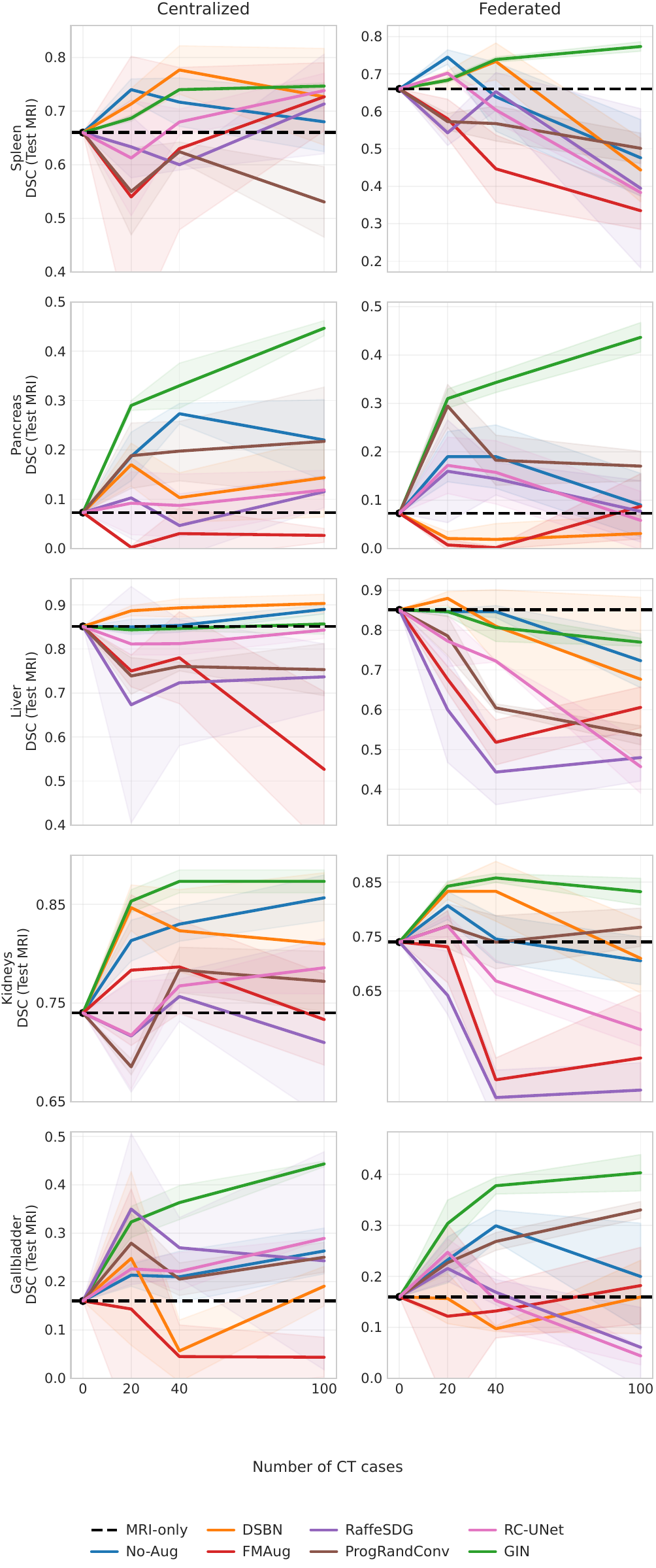}
   \caption{
Cross-modality medical image segmentation performance on MRI test data using CT–MRI training under different CT ratios.
Results are shown for five organs (Spleen, Pancreas, Liver, Kidneys, and Gallbladder), one per row.
The left column reports centralized training, while the right column reports federated training.
Solid curves denote mean Dice scores across three runs, and shaded regions indicate $\pm$ one standard deviation.
The dashed horizontal line indicates the MRI-only baseline.
}
    \label{fig:results}
\end{figure*}

\subsubsection{Results}

\paragraph{Limited Data Collaboration}
Figure~\ref{fig:results} and Table~\ref{tab:organ_limited_summary} summarize the impact of progressively incorporating CT data into MRI-limited training. The benefits of cross-modality collaboration are highly organ-dependent. For small and anatomically complex structures such as the gallbladder and pancreas, performance gains are substantial: +151.9\% and +498.6\%, respectively. In contrast, the liver, which already achieves a strong baseline MRI-only (DSC = 0.851), shows only marginal improvement in centralized training and even some drop in performance in FL. This pattern reveals that for the gallbladder and pancreas, the MRI-only baseline completely fails (DSC < 0.30), and cross-modality collaboration becomes essential rather than merely beneficial, enabling recovery from complete failure to functional performance through knowledge transfer from CT. For segmentation challenging organs, performance improves monotonically as more CT cases are added ($20 \rightarrow 40 \rightarrow 100$), with the pancreas progressing from near-failure (DSC = 0.073) to functional performance (DSC = 0.437), illustrating effective knowledge transfer across modalities. Improvement patterns were similar between GIN-augmented FL and central training across all organs, except the liver.

\paragraph{Cross-Modality Generalization}
Table~\ref{tab:organ_results} presents the full cross-modality evaluation results. Models trained on a single modality fail substantially when tested on the other—MRI-only models fail on CT, and CT-only models fail on MRI, confirming the lack of generalizability in unimodal training. Integrating multimodal data from multiple institutions, our FL approach consistently achieves strong performance across both CT and MRI. Notably, FedGIN retains 93--98\% of centralized GIN performance, with only minor drops (1-3\%) for most organs, demonstrating that effective cross-modality generalization is achievable without data centralization.

Among all methods, GIN consistently outperforms network-level adaptation (DSBN), frequency-domain approaches (FMAug, RaffeSDG), and spatial-domain alternatives (ProRandConv, RC-Unet), in both centralized and federated settings. Other augmentation methods fail substantially in the federated setting despite performing adequately in centralized evaluation: DSBN collapses on MRI (DSC dropping to 0.001-0.14), and frequency-domain methods show similar instability. This degradation is attributed to two key factors: DSBN's domain-specific batch statistics conflict with federated model aggregation, and frequency-domain perturbations prove overly sensitive to the intensity variations present across clients. GIN's spatial-domain approach, by contrast, applies simple random convolutional transformations locally before aggregation, generating stable augmented samples that remain robust across federated optimization. The practical implication is that a hospital could participate in a federated network where different sites have different scanners or modalities, resulting in a single model that works well across both modalities, even if each institution contributed data from only one imaging type

\subsection{Use case 2: Whole Heart Segmentation}
\label{2dwhs}

\subsubsection{Dataset}

We utilize the CARE-Whole Heart Segmentation Challenge 2025 dataset~\cite{zhang2018translating} for multi-center whole heart segmentation. This dataset provides cardiac CT and MRI volumes with annotations for seven cardiac substructures: left ventricular blood cavity (LV), right ventricular blood cavity (RV), left atrial blood cavity (LA), right atrial blood cavity (RA), myocardium of the left ventricle (Myo), ascending aorta (AO), and pulmonary artery (PA). CT volumes have an in-plane resolution of approximately $0.78 \times 0.78$~mm with slice thickness of 1.6~mm, while MRI volumes have approximately 1.0~mm isotropic resolution. 

Our multi-center federated setup comprises 86 3D volumes distributed across four clients, as summarized in Table~\ref{tab:heart_data}. Each client accesses only its local data and modality. Client 4 is held out exclusively for evaluation at the federated server, with its 26 cases split equally: 13 cases are used for validation during training for model selection, and the remaining 13 cases serve as the final unseen test set for performance reporting.

\begin{table}[h]
\centering
\caption{Distribution of 3D volumes across centers for whole heart segmentation from CARE Challenge 2025 dataset.}
\label{tab:heart_data}
\begin{tabular}{lccll}
\hline
\textbf{Split} & \textbf{Client} & \textbf{Center} & \textbf{Modality} & \textbf{Volumes} \\
\hline
\multirow{3}{*}{Training} & Client 1 & Center A & CT & 20 \\
                          & Client 2 & Center B & CT & 20 \\
                          & Client 3 & Center C \& D & MRI & 20 \\
\hline
Validation/Testing & Client 4 & Center E & MRI & 26 \\
\hline
\end{tabular}
\end{table}

\begin{sidewaystable*}
\centering
\caption{Cross-Modality Whole Heart Segmentation Performance (DSC) on Client 4 MRI Test Data. The MRI-only baseline is trained on Client 3 (20 cases). Cross-modality collaboration uses Client 3 (MRI) and Clients 1\&2 (CT). Abbreviations: Myo = Myocardium, LA = Left Atrium, LV = Left Ventricle, RA = Right Atrium, RV = Right Ventricle, AO = Aorta, PA = Pulmonary Artery. Best results in \textbf{bold}, second best \underline{underlined} for Centralized and Federated training strategies separately.}
\label{tab:whs_results}
\small
\begin{tabular}{llcccccccc}
\hline
\textbf{Training Strategy} & \textbf{Method} & \textbf{Myo} & \textbf{LA} & \textbf{LV} & \textbf{RA} & \textbf{RV} & \textbf{AO} & \textbf{PA} & \textbf{Mean} \\
\hline
Local & MRI-only & 0.5936 & 0.7178 & 0.7376 & 0.5971 & 0.4923 & 0.4618 & 0.3930 & 0.5705 \\
\hline
\multirow{3}{*}{Centralized} 
& No augmentation & \underline{0.6352} & \textbf{0.7347} & \underline{0.7577} & \underline{0.6943} & \underline{0.5894} & \underline{0.6439} & \underline{0.5208} & \underline{0.6537} \\
& ProRandConv & 0.3986 & 0.2145 & 0.6981 & 0.4565 & 0.3968 & 0.6117 & 0.5319 & 0.4726 \\
& GIN & \textbf{0.7088} & \underline{0.6647} & \textbf{0.8749} & \textbf{0.6709} & \textbf{0.6602} & \textbf{0.6169} & \textbf{0.5483} & \textbf{0.6778} \\
\hline
\multirow{3}{*}{Federated} 
& FedAvg & \underline{0.3807} & \textbf{0.6369} & \underline{0.6974} & \textbf{0.6528} & \underline{0.4626} & \underline{0.6078} & \underline{0.5080} & \underline{0.5637} \\
& ProRandConv & 0.3915 & 0.2740 & 0.7413 & 0.4587 & 0.2126 & 0.5787 & 0.3538 & 0.4301 \\
& FedGIN & \textbf{0.6379} & \underline{0.5347} & \textbf{0.8643} & \underline{0.6327} & \textbf{0.6569} & \textbf{0.6380} & \textbf{0.4435} & \textbf{0.6297} \\
\hline
\hline
\multicolumn{2}{l}{\textbf{Improvement (FedGIN vs MRI-only)}} & \textbf{+7.46\%} & \textbf{-25.49\%} & \textbf{+17.17\%} & \textbf{+5.96\%} & \textbf{+33.44\%} & \textbf{+38.16\%} & \textbf{+12.85\%} & \textbf{+10.37\%} \\
\hline
\end{tabular}
\end{sidewaystable*}

\subsubsection{Experimental Setup and Motivation}

This task evaluates multi-class segmentation (7 cardiac substructures) in a multi-center federated scenario, we investigate whether MRI-limited centers (Client 3) can benefit from collaboration with CT-equipped centers (Clients 1\&2) through FL. The model is evaluated on the held-out Client 4 MRI data, simulating deployment at a center that did not participate in training.
For comparison, we establish two baselines: (1) an MRI-only model trained solely on Client 3's local data, and (2) a centralized model trained on all available data from Clients 1-3 (60 volumes combining CT and MRI modalities). 
Based on the findings from Use case 1, we focus our comparison on the two strongest augmentation strategies: GIN and ProRandConv. The federated configuration uses $K=3$ clients (two CT clients and one MRI client), performing multi-class segmentation across seven cardiac structures.

\subsubsection{Results}

Table~\ref{tab:whs_results} reports segmentation performance on the held-out Center E MRI test data. FedGIN achieves a mean DSC of 0.6297, an absolute gain of 0.0592 over the MRI-only baseline, confirming that cross-modality collaboration improves multi-class cardiac segmentation under realistic multi-center domain shifts.

Performance gains are structure-dependent. RV, LV and AO show the largest improvements, while LA performance decreases relative to baseline. This pattern suggests that thick-walled structures benefit more from cross-modality augmentation than thin-walled atria, pointing to structure-specific cross-modality robustness. FedGIN retains 93\% of centralized GIN performance (0.6297 vs. 0.6778), showing that FL achieves slightly lower performance than centralized training. GIN consistently outperforms ProRandConv in both centralized (0.6778 vs. 0.4726) and federated (0.6297 vs. 0.4301) settings, validating its robustness in multi-class segmentation tasks.

\subsubsection{Native 3D Whole-Heart Segmentation with nnU-Net}
\label{sec:3d-whs}

Unlike the 2D whole-heart experiments in Section~\ref{2dwhs}, which use the CARE-WHS 2025 dataset, the native 3D experiments described here use the CARE-WHS 2026 dataset \footnote{\url{https://zmic.org.cn/care_2026/track_wholeheart/}}. To evaluate whether the cross-modality federated behavior observed in the 2D experiments persists under native volumetric segmentation, we conduct a complementary 3D study using this dataset. We adopt a self-configuring nnU-Net architecture~\cite{isensee2021nnu}, which automatically determines preprocessing, patch size, and network topology from the training data. Unlike the fixed 2D U-Net backbone used elsewhere in this study, nnU-Net operates directly on full 3D volumes rather than axial slices, removing the need for slice-wise reconstruction at evaluation time. The GIN augmentation module is adapted to this setting by replacing the 2D random convolutions with their 3D counterparts (kernel sizes $1\times1\times1$ and $3\times3\times3$), as described in Section~\ref{meth}. Centralized models are trained for 1000 epochs, following the standard nnU-Net training schedule. For the FL experiments, training runs for $T=1000$ communication rounds with $E=1$ local epoch per round, matching the total local update budget of centralized training.

\begin{table}[h]
\centering
\caption{Distribution of 3D volumes across centers for whole heart segmentation from the CARE-WHS 2026 dataset (106 cases total).}
\label{tab:heart_data_3d}
\begin{tabular}{lccc}
\hline
\textbf{Split} & \textbf{Center} & \textbf{Modality} & \textbf{Num. Patients} \\
\hline
\multirow{3}{*}{Training} & Center A & CT  & 20 \\
                          & Center B & CT  & 20 \\
                          & Center C \& D & MRI & 20 \\
\hline
\multirow{2}{*}{Held-out (Testing)} & Center G & CT  & 20 \\
                                    & Center E & MRI & 26 \\
\hline
\textbf{Total} & & & \textbf{106} \\
\hline
\end{tabular}
\end{table}

The federated configuration comprises three clients: two CT-only clients (Centers A and B) and one MRI-only client (Center C\&D), following the same single-modality-per-client constraint as the 2D experiments. Table~\ref{tab:heart_data_3d} summarizes the entire cohort of CARE-WHS 2026. Generalization is evaluated in two centers that were not seen during training, Center G (CT) and Center E (MRI), enabling a bidirectional assessment of cross-modality retention, in contrast to the unidirectional (MRI-only) holdout used in the 2D whole-heart evaluation. Center E, held out for MRI evaluation here, is the same center used as the sole held-out MRI test site in the 2D whole-heart evaluation (Section~\ref{2dwhs}), as the CARE-WHS challenge is released as a continuing annual series that retains its core sites (Centers A, B, C\&D, and E) across editions while introducing additional sites in later releases. The CARE-WHS 2026 release used in this section adds Center G, a newly introduced CT site, enabling the bidirectional CT/MRI holdout evaluation that is not possible with the 2025 release alone.

We compare five configurations: (1) three local models, each trained solely on one client's data; (2) centralized, no augmentation, trained on pooled CT+MRI data; (3) centralized GIN; (4) federated, no augmentation (FedAvg); and (5) federated GIN (FedGIN). To confirm whether the gains observed with GIN in the primary 2D experiments also extend to volumetric segmentation, we conduct a targeted 3D whole-heart experiment using a self-configuring nnU-Net architecture. The 3D evaluation therefore focuses on comparing GIN against the corresponding no-augmentation baseline, while a broader comparison of additional augmentation and adaptation strategies is left for future work.

\begin{sidewaystable*}
\centering
\caption{Cross-Modality Whole Heart Segmentation Performance (DSC) on the Unseen CT Center (Center G), Native 3D nnU-Net, CARE-WHS 2026. Local rows report each single-modality model evaluated on Center G. Best result between the two methods within each of Centralized and Federated shown in \textbf{bold}.}
\label{tab:whs_3d_ct_results}
\begin{tabular}{llcccccccc}
\hline
\textbf{Training Strategy} & \textbf{Method} & \textbf{Myo} & \textbf{LA} & \textbf{LV} & \textbf{RA} & \textbf{RV} & \textbf{AO} & \textbf{PA} & \textbf{Mean} \\
\hline
\multirow{3}{*}{Local}
& CT-A only   & 0.9602 & 0.9628 & 0.9642 & 0.9433 & 0.9413 & 0.9793 & 0.9110 & 0.9517 \\
& CT-B only   & 0.9709 & 0.9806 & 0.9703 & 0.9534 & 0.9548 & 0.9840 & 0.9427 & 0.9652 \\
& MRI-CD only & 0.4323 & 0.0722 & 0.5418 & 0.1037 & 0.2337 & 0.6227 & 0.3280 & 0.3335 \\
\hline
\multirow{2}{*}{Centralized}
& No augmentation & \textbf{0.9689} & \textbf{0.9798} & \textbf{0.9701} & 0.9476 & \textbf{0.9560} & 0.9799 & \textbf{0.9470} & \textbf{0.9642} \\
& GIN             & 0.9613 & 0.9781 & 0.9659 & \textbf{0.9597} & 0.9494 & \textbf{0.9813} & 0.9450 & 0.9630 \\
\hline
\multirow{2}{*}{Federated}
& FedAvg  & 0.9249 & 0.9608 & 0.9511 & 0.7594 & 0.8631 & 0.9350 & 0.6930 & 0.8696 \\
& FedGIN  & \textbf{0.9432} & \textbf{0.9622} & \textbf{0.9544} & \textbf{0.8643} & \textbf{0.8712} & \textbf{0.9357} & \textbf{0.6996} & \textbf{0.8901} \\
\hline
\end{tabular}

\vspace{2em}

\captionof{table}{Cross-Modality Whole Heart Segmentation Performance (DSC) on the Unseen MRI Center (Center E), Native 3D nnU-Net, CARE-WHS 2026. Local rows report each single-modality model evaluated on Center E. Best result between the two methods within each of Centralized and Federated shown in \textbf{bold}.}
\label{tab:whs_3d_mri_results}
\begin{tabular}{llcccccccc}
\hline
\textbf{Training Strategy} & \textbf{Method} & \textbf{Myo} & \textbf{LA} & \textbf{LV} & \textbf{RA} & \textbf{RV} & \textbf{AO} & \textbf{PA} & \textbf{Mean} \\
\hline
\multirow{3}{*}{Local}
& CT-A only   & 0.1677 & 0.0724 & 0.1931 & 0.0012 & 0.0528 & 0.0396 & 0.0142 & 0.0773 \\
& CT-B only   & 0.4067 & 0.4278 & 0.5587 & 0.0951 & 0.2877 & 0.3218 & 0.0567 & 0.3078 \\
& MRI-CD only & 0.7955 & 0.8424 & 0.9342 & 0.8106 & 0.7490 & 0.5171 & 0.4950 & 0.7348 \\
\hline
\multirow{2}{*}{Centralized}
& No augmentation & 0.7493 & 0.8076 & 0.8815 & 0.7921 & 0.7202 & 0.5231 & 0.5014 & 0.7107 \\
& GIN             & \textbf{0.8238} & \textbf{0.8972} & \textbf{0.9344} & \textbf{0.8860} & \textbf{0.8591} & \textbf{0.6067} & \textbf{0.5490} & \textbf{0.7937} \\
\hline
\multirow{2}{*}{Federated}
& FedAvg  & 0.7029 & 0.8312 & 0.9169 & 0.7713 & 0.6883 & 0.5901 & 0.5111 & 0.7160 \\
& FedGIN  & \textbf{0.8255} & \textbf{0.8895} & \textbf{0.9431} & \textbf{0.8936} & \textbf{0.8405} & \textbf{0.6337} & \textbf{0.5436} & \textbf{0.7956} \\
\hline
\end{tabular}
\end{sidewaystable*}

Tables~\ref{tab:whs_3d_ct_results} and~\ref{tab:whs_3d_mri_results} report per-structure and overall DSC on the two held-out centers. As expected, models trained on a single modality generalize poorly to the other, CT-only local models achieve overall DSC of 0.33--0.31 on the unseen MRI center, and the MRI-only local model achieves 0.33 on the unseen CT center, confirming the same failure mode observed in the 2D experiments (Section~\ref{2dwhs}). Centralized training without augmentation performs strongly on CT (0.9642) but generalizes weakly to MRI (0.7107); adding GIN augmentation improves MRI generalization substantially (0.7107 $\rightarrow$ 0.7937) while leaving CT performance essentially unchanged (0.9642 $\rightarrow$ 0.9630), mirroring the 2D pattern in which GIN's benefit is concentrated on the modality most affected by domain shift. Notably, GIN's centralized CT improvement is not uniform across structures: it improves RA and AO over no-augmentation, but is marginally lower on Myo, LA, LV, RV, and PA, indicating a modest trade-off rather than a uniform gain.

Under federated setting, GIN outperforms the no-augmentation baseline on every structure and both held-out centers: CT mean DSC improves from 0.8696 to 0.8901 and MRI mean DSC improves marginally from 0.7160 to 0.7956, with particularly large gains on the previously weak RA and RV structures under CT (0.7594 $\rightarrow$ 0.8643 and 0.8631 $\rightarrow$ 0.8712, respectively). Relative to centralized GIN, FedGIN retains 92.4\% of mean performance on the unseen CT center, consistent with the 93--98\% retention range observed in the 2D use cases. On the unseen MRI center, however, FedGIN (0.7956) is essentially at parity with,  marginally exceeding, centralized GIN (0.7937). The 3D WHS results are broadly consistent with the primary 2D findings, with GIN improving FL performance over the no-augmentation baseline on both held-out CT and MRI data. This confirms that the cross-modality generalization benefit of GIN also extends to native volumetric segmentation using a self-configuring nnU-Net architecture. Differences in the degree of performance retention across modalities are reported descriptively, without assigning a specific cause, since the 3D study was designed as a targeted confirmatory experiment with a single run per configuration. Across both tables, the AO and PA remain the weakest-performing structures in nearly every configuration on MRI, consistent with a persistent difficulty in segmenting these thin vascular structures that augmentation alone does not resolve. Establishing whether the CT/MRI asymmetry in federated retention is a robust effect, and addressing the AO/PA gap, are directions we are pursuing in a dedicated follow-up study using this dataset.

\subsection{Discussion}

Across both tasks and all experimental configurations, GIN consistently outperforms frequency-domain augmentation, network-level adaptation, and other spatial-domain methods. This robustness can be attributed to GIN's mechanism of applying random convolutional transformations locally before federated aggregation, which generates diverse and realistic intensity variations that simulate cross-modality appearance differences while preserving anatomical structure. Unlike frequency-domain methods, which are sensitive to cross-client distribution shifts during aggregation, and DSBN, whose modality-specific statistics do not compose well under federated averaging, GIN's spatial-domain approach remains stable across the heterogeneous optimization landscape of FL.

\textbf{Limitations and scope:} 
Our experiments deliberately operate in a low-data regime (20-100 training volumes per modality) to evaluate cross-modality learning when local data is scarce. While many modern datasets contain thousands of patient examinations, institutional silos in federated settings often face data scarcity, particularly for specialized imaging modalities or rare conditions. However, our absolute performance numbers remain substantially below state-of-the-art centralized models: TotalSegmentator achieves Dice Score (DSC) of 82.8\% for pancreas and 92.5\% for gallbladder on CT, compared to our best federated results of 43.7\% and 59.9\%, respectively. This gap reflects both our limited training data and the inherent challenge of cross-modality generalization without paired data. Additionally, our federated experiments involve only 2-3 clients, whereas real-world multi-institutional collaboration could involve more hospitals, which may introduce additional communication and heterogeneity challenges. The 3D experiments (Section~\ref{sec:3d-whs}) carry additional limitations specific to that setting, discussed there in detail.

\textbf{Future directions:} Several avenues remain open for investigation. First, we currently use all augmented images for training, but some may be low-quality or unrealistic. Filtering out poor augmentations could improve model performance \cite{vsvabensky2025evaluating}, although defining “quality” in medical image enhancement remains an open question. Second, our primary experiments focused on the 2D U-Net backbone; Section~\ref{sec:3d-whs} presents an initial extension to nnU-Net~\cite{isensee2021nnu} (the backbone of TotalSegmentator), though a systematic comparison across augmentation strategies on this architecture remains open. Extending FedGIN to foundation models such as SAM and MedSAM~\cite{MedSAM}, or recent vision transformers, represents a further direction. Third, we present an initial 3D whole-heart evaluation (Section~\ref{sec:3d-whs}) confirming that GIN-based federated augmentation extends to native volumetric segmentation. The 3D WHS experiment serves as a targeted confirmatory evaluation of the findings observed in the primary 2D study. Using a self-configuring nnU-Net architecture, GIN consistently improves performance over the corresponding no-augmentation baseline on held-out CT and MRI data, providing evidence that the cross-modality generalization benefit extends beyond slice-based 2D segmentation to a strong native 3D segmentation setting. A broader evaluation across additional 3D settings will be considered in future work.

Looking across all experiments, our results show that cross-modality FL is effective, especially when baseline performance is low. GIN augmentation consistently outperforms other methods, with its benefits most pronounced in federated settings, where other methods tend to degrade. FL retains the centralized performance across both 2D and 3D evaluations, typically 92--98\%, with federated MRI performance in the 3D setting occasionally matching or marginally exceeding centralized GIN, making it a viable, privacy-preserving alternative for multi-institutional collaboration. Crucially, we show that effective cross-modality models can be trained without centralizing data, enabling hospitals with diverse imaging equipment to collaborate while keeping patient data local.

\section{Conclusion}
\label{conc}
This work presents a FL framework for cross-modality medical image segmentation that enables collaboration between institutions using unpaired CT and MRI data, without requiring data centralization. Our experiments on abdominal organ and whole-heart segmentation highlight three key findings: (1) significant gains in low-data settings and for segmentation-challenging organs, (2) strong generalization across modalities while closely approximating centralized performance, and (3) native 3D experiments confirming that these cross-modality FL benefits extend beyond slice-based 2D training. Overall, this work demonstrates that federated cross-modality learning can achieve competitive segmentation performance compared to centralized training, while preserving data privacy and enabling practical multi-institutional AI collaboration across diverse healthcare systems. Future work will extend this evaluation to a broader set of augmentation strategies and architectures under native 3D training, and to larger, more heterogeneous federations.

\section*{Declarations}

\bmhead{Funding}
This work was supported by the Norwegian Cancer Society through the FLIP.AI project (Federated Learning to Improve Prostate cancer imaging with Artificial Intelligence), Project ID 2542041, Norwegian University of Science and Technology, hosted by the Department of Circulation and Medical Imaging, Faculty of Medicine and Health Sciences.

\bmhead{Conflict of interest/Competing interests}
The authors declare that they have no competing interests.

\bmhead{Ethics approval and consent to participate}
Not applicable. This study used only publicly available, de-identified datasets (TotalSegmentator, AMOS2020) and challenge-provided, de-identified datasets (CARE-WHS 2025, CARE-WHS 2026) released for research use by their respective organizers; no new human participant data were collected by the authors.

\bmhead{Consent for publication}
Not applicable.

\bmhead{Data availability}
The TotalSegmentator and AMOS2020 datasets used for abdominal organ segmentation are publicly available at \url{[https://github.com/wasserth/totalsegmentator]} and \url{[https://zenodo.org/records/7262581]}, respectively. The CARE-WHS 2025 and CARE-WHS 2026 datasets used for whole-heart segmentation are not publicly available; access is restricted to registered participants of the respective CARE-WHS challenges and is subject to the challenge organizers' data use agreements. Requests for access should be directed to the CARE-WHS challenge organizers.

\bmhead{Code availability}
The code used in this study is publicly available at \url{https://github.com/sachugowda/FedGINwithExtension}.




\nocite{*}
\bibliography{sn-bibliography}

\end{document}